\documentclass[lettersize,journal]{IEEEtran}
\usepackage{amsmath,amsfonts}
\usepackage{algorithmic}
\usepackage{algorithm}
\usepackage{array}
\usepackage{textcomp}
\usepackage{stfloats}
\usepackage{url}
\usepackage{verbatim}
\usepackage{graphicx}
\usepackage{cite}
\usepackage{orcidlink}
\usepackage{hyperref}
\usepackage{url}
\usepackage{tikz}
\usetikzlibrary{calc}

\usepackage{subcaption}
\usepackage{array} 
\usepackage{multirow}
\usepackage{booktabs} 
\usepackage{adjustbox}
\usepackage{makecell}
\usepackage{enumitem}
\usepackage{bm}
\usepackage[fleqn]{nccmath}
\usepackage{amsthm}
\usepackage{xparse,amssymb}

\newtheorem{theorem}{Theorem}

\newtheorem{mydef}[theorem]{Definition}

\newtheorem{myremark}[theorem]{Remark}

\DeclareMathOperator*{\prox}{\text{prox}}

\newcommand{\R}{\mathbb{R}}

\begin{document}

\title{A multiobjective continuation method to compute the regularization path of deep neural networks}

\author{Augustina C. Amakor\,\orcidlink{0009-0000-3648-6901}, Konstantin Sonntag\,\orcidlink{0000-0003-3384-3496} and Sebastian Peitz \,\orcidlink{0000-0002-3389-793X}}




\maketitle

\begin{abstract}
Sparsity is a highly desired feature in deep neural networks (DNNs) since it ensures numerical efficiency, improves the interpretability of models (due to the smaller number of relevant features), and robustness. For linear models, it is well known that there exists a \emph{regularization path} connecting the sparsest solution in terms of the $\ell^1$ norm, i.e., zero weights and the non-regularized solution. Very recently, there was a first attempt to extend the concept of regularization paths to DNNs by means of treating the empirical loss and sparsity ($\ell^1$ norm) as two conflicting criteria and solving the resulting multiobjective optimization problem for low-dimensional DNN. However, due to the non-smoothness of the $\ell^1$ norm and the high number of parameters, this approach is not very efficient from a computational perspective for high-dimensional DNNs. To overcome this limitation, we present an algorithm that allows for the approximation of the entire Pareto front for the above-mentioned objectives in a very efficient manner for high-dimensional DNNs with millions of parameters. We present numerical examples using both deterministic and stochastic gradients. We furthermore demonstrate that knowledge of the regularization path allows for a well-generalizing network parametrization.
To the best of our knowledge, this is the first algorithm to compute the regularization path for non-convex multiobjective optimization problems (MOPs) with millions of degrees of freedom.
\end{abstract}


\section{Introduction}
Machine Learning (ML) and in particular deep neural networks (DNNs) are nowadays an integral part of numerous applications such as data classification, image recognition, time series prediction, and language processing. Their importance continues to grow at great speed across numerous disciplines and applications, and the increase in available computational capacity allows for the construction of larger models. However, these advances also increase the challenges regarding the construction and training of DNNs, e.g., the required training data, the training efficiency, and the adaptability to changing external factors. This leads to the task of simultaneously fulfilling numerous, sometimes contradictory goals in the best possible way.

Multiobjective optimization (MO) addresses the problem of optimizing several conflicting objectives. The issue of having to trade between multiple, conflicting criteria is a universal problem, such as the need to have an optimal tradeoff between cost and quality in a production process. In a similar manner, conflicting criteria occur in various ways in ML and can be addressed as MOPs which makes multiobjective optimization very important in ML. The main task is thus to identify the set of optimal trade-off solutions (the \emph{Pareto set}) between these conflicting criteria. This concerns multitask problems \cite{Sener2018}, but also the training itself, where we want to minimize the training loss, obtain sparse models and improve generalization. 

Interestingly, we found that many papers on multicriteria machine learning do not address the true nature of multiobjective optimization. For instance, when choosing very large neural network architectures or even considering task-specific layers \cite{Sener2018}, the different tasks (objectives) do not conflict. The network is simply too powerful such that both tasks can be optimally handled simultaneously and there is no tradeoff. From an optimization point of view, the Pareto front collapses into a single point. Also, low-dimensional problems are often times considered such as in the case of evolutionary algorithms \cite{zitzler1998, Deb2001, Zhang2007, Deb2011, Blank2020} which are usually computationally expensive. 
However, considering the strongly growing carbon footprint of AI \cite{Gibney2022}, there is a growing interest in smaller models that are better tailored to specific tasks. This is why we propose the use of models that are smaller and adapted to a certain task for high-dimensional DNNs. While this will reduce the general applicability, multicriteria optimization can help us to determine a set of compromising network architectures, such that we can adapt networks to specific situations online (multicriteria decision making). Through our approach, we show the first algorithm that solves a truly high-dimensional deep learning problem in a very efficient manner.

The joint consideration of loss and $\ell^1$ regularization is well-studied for linear systems. However, it is much less understood for the nonlinear problems that we face in deep learning. In DNN training, the regularization path is usually not of interest. Instead, methods aim to find a single, suitable trade-off between loss and $\ell^1$ norm \cite{chen2021, bregman2022, fu2020, fu2022, lemhadri2021}. When interpreting the $\ell^1$ regularization problem as a multiobjective optimization problem (MOP), a popular approach to obtain the entire solution set (the \emph{Pareto set}) is via \emph{continuation methods} \cite{Hillermeier2001,SDD2005}. They usually consist of a predictor step (along the tangent direction of the Pareto set) and a corrector step that converges to a new point on the Pareto set close by. However, as the $\ell^1$ norm is non-smooth, classical manifold continuation techniques fail. Due to this fact, a first extension of regularization paths from linear to nonlinear models was recently presented in \cite{Bieker2022}, where continuation methods were extended to non-smooth objective functions. Although this extension provides a rigorous treatment of the problem, it results in a computationally expensive algorithm, which renders it impractical for DNNs of realistic dimensions.

As research on regularization paths for nonlinear loss functions has focused on small-scale learning problems until now, this work is concerned with large-scale machine learning problems. The main contributions of this paper include:
\begin{itemize}
\item The extension of regularization paths from linear models to high-dimensional nonlinear deep learning problems.
\item The demonstration of the usefulness of multiobjective continuation methods for the generalization properties of DNNs.
\item A step towards more resource-efficient ML by beginning with very sparse networks and slowly increasing the number of weights. This is in complete contrast to the standard pruning approaches, where we start very large and then reduce the number of weights.
\end{itemize}
A comparison of our approach is further made with the standard approach of weighting the individual losses using an additional hyperparameter, i.e., the weighted sum method, and with one of the most popular evolutionary algorithms, the non-dominated sorting genetic algorithm (NSGA-II).

The remainder of the paper is organized as follows. In Section 2 we discuss the various related works. In Section 3, we present our continuation method together with the algorithms. Section 4 presents a detailed discussion of our extensive numerical experiments and results. Finally, conclusions are drawn in Section 5.

\section{Related works}
\label{Relate}

\subsection{Multiobjective optimization}
In the last decades, many approaches have been introduced for solving nonlinear \cite{Miettinen1998}, non-smooth \cite{Poirion2017}, non-convex \cite{Miettinen1998}, or stochastic \cite{Mitrevski2020} multiobjective optimization problems, to name just a few special problem classes. In recent years, researchers have further begun to consider multiobjective optimization in machine learning \cite{Qu2021, Jin2008, Deb2011, Zhang2007} and deep learning \cite{Sener2018, Ruchte2021}. We provide an overview of the work that is most pertinent to our own and direct readers to the works \cite{Sener2018} and \cite{Bieker2022}. 

Reference \cite{Ehrgott2008} described MO as the process of optimizing multiple, sometimes conflicting objectives simultaneously by finding the set of points that are \textit{not dominated}\footnote{A point is said to \emph{dominate} another point if it performs at least as good as the other point across all objectives and strictly better than it in at least one objective.} by any other feasible points. 
Different methods have been proposed to obtain Pareto optimal solutions such as \textit{evolutionary algorithms} which use the evolutionary principles inspired by nature by evolving an entire population of solutions \cite{Deb2001}. \textit{Gradient-based techniques} extend the well-known gradient techniques from single objective optimization to multiple criteria problems \cite{Peitz2018}. \textit{Set oriented methods} provide an approximation of the Pareto set through box coverings and often suffer from the curse of dimensionality which makes applications in ML very expensive \cite{Dellnitz2005}. Another deterministic approach are \textit{scalarization methods} which involve the transformation of MOP into (a series of) single objective optimization problems \cite{Eichfelder2008}. These can then be solved using standard optimization techniques. Examples include the epsilon constraint, Chebyshev scalarization, goal programming, min-max, augmented epsilon constraint and weighted sum method \cite{Ehrgott2008},\cite{Kim2006}. Some drawbacks exist for the latter methods, most notably the incapability of some to handle non-convex problems, as well as the challenge to obtain equidistant coverings. Notably, these drawbacks are most severe in the weighted sum method, which is by far the most widely applied approach in ML when considering multiple criteria at the same time (such as regularization terms). Moreover, the weighting has to be made a priori, which makes selecting useful trade-offs much harder \cite{Ehrgott2008, Sener2018}.

\textit{Continuation methods} are another approach for the computation of the solutions of MOPs. These methods were initially developed to solve complex problems starting from simple ones, using homotopy approaches\footnote{These are approaches that involve starting at a simple-to-calculate solution and then continuously vary some parameter to increase the problem difficulty step by step, until finally arriving at the solution of the original problem, which is often very hard to compute directly \cite{Forster1995}.}. Also, they were used in predictor-corrector form to track manifolds \cite{Allgower1990, Chow1991}. In the multiobjective setting, one can show that the Pareto set is a manifold as well\footnote{To be more precise, the set of points satisfying the Karush-Kuhn-Tucker (KKT) necessary conditions for optimality along with the corresponding KKT multipliers form a manifold of dimension $m-1$.}, if the objectives are sufficiently regular \cite{Hillermeier2001}. Hence, the continuation method then becomes a technique to follow along the Pareto set in two steps; \textbf{a)} a predictor step along the tangent space of the Pareto set which forms in the smooth setting a manifold of dimension \textit{m - 1} (where $m$ is the number of objective functions) \cite{Hillermeier2001} and \textbf{b)} a corrector step that obtains the next point on the Pareto set which is often achieved using multiobjective gradient descent. \cite{SDD2005} used the predictor-corrector approach to find points that satisfy the Karush-Kuhn-Tucker (KKT) condition and to further identify other KKT points in the neighbourhood.

The continuation method has further been extended to regularization paths in machine learning (ML) for linear models \cite{Park2007, Guha2023}. Computing the regularization path for nonlinear models by solving a multiobjective optimization problem has been introduced recently although limited to small dimensions since dealing with non-smoothness is difficult \cite{Bieker2022}.

\subsection{Multicriteria machine learning}

In the context of multicriteria machine learning, several approaches have been used such as \textit{evolutionary algorithms} \cite{Deb2002, Bernadi2001, Jin2008} or \textit{gradient-based methods} \cite{Sener2018, Mitrevski2020}. However, only a few of these methods address high-dimensional deep learning problems or attempt to compute the entire Pareto front. Furthermore, as discussed in the introduction, many researchers have introduced architectures that are so powerful that the Pareto front collapses into a single point (e.g., \cite{Sener2018}). In the same way, the regularization path is usually not of much interest in DNN training. Instead, most algorithms yield a single trade-off solution for loss and $\ell^1$ norm that is usually influenced by a weighting parameter \cite{chen2021, bregman2022, fu2022, lemhadri2021}. However, we want to pursue a more sustainable path with well-distributed solutions and we are interested in truly conflicting criteria. The entire regularization path for DNNs (i.e., a MOP with training loss versus $\ell^1$ norm) was computed in \cite{Bieker2022}. However, even though the algorithm provably yields the Pareto front, the computation becomes intractable in very high dimensions. Hence, the need to develop an efficient method to find the entire regularization path and Pareto front for DNNs.

For high-dimensional problems, gradient-based methods have proven to be the most efficient. Examples are the steepest descent method \cite{Fliege2000}, projected gradient method \cite{Drummond2004}, proximal gradient method \cite{Tanabe2019, chen2021} and accelerated proximal gradient method \cite{Tanabe2022, Sonntag2022}. Previous approaches for MOPs often assume differentiability of the objective functions but the $\ell^1$ norm is not differentiable, so we use the multiobjective proximal gradient method (MPG) to ensure convergence. MPG has been described by \cite{Tanabe2019} as a descent method for unconstrained MOPs where each objective function can be written as the sum of a convex and smooth function, and a proper, convex, and lower semicontinuous but not necessarily differentiable one. Simply put, MPG combines the proximal point and the steepest descent method and is a generalization of the iterative shrinkage-thresholding algorithm (ISTA) \cite{Comb2005, Beck2009} to multiobjective optimization problems.  

\section{Continuation Method}

\subsection{Some basics on multiobjective optimization}
A multiobjective optimization problem can be mathematically formalized as 
\begin{align}
        \min_{\theta \in \R^n} \left[ \begin{array}{c}
            F_1(\theta)  \\
            \vdots \\
            F_m(\theta) 
        \end{array}\right],
    \label{eq:MOP}
    \tag{MOP}
\end{align}
where $ F_i : \mathbb{R}^n \rightarrow \mathbb{R}$ for $i = 1, \dots, m $ are the objective functions and $\theta$ the parameters we are optimizing. In general, there does not exist a single point that minimizes all criteria simultaneously. Therefore, the solution to \eqref{eq:MOP} is defined as the so-called \emph{Pareto set}, which consists of optimal compromises. 

\begin{mydef} Consider the optimization problem \eqref{eq:MOP}.
 \begin{enumerate}
 \item A point $ \theta^* \in \R^n$ is \emph{Pareto optimal} if there does not exist another point $ \theta \in \R^n$ such that $ F_i(\theta) \le F_i(\theta^*)$ for all $i = 1,\dots,m,$ and $F_j(\theta) < F_j(\theta^*)$ for at least one index $j$. The set of all Pareto optimal points is the \emph{Pareto set}, which we denote by $P$. The set $F(P) \subset \R^m$ in the image space is called the \emph{Pareto front}.

 \item  A point $\theta^* \in \R^n$ is said to be \emph{weakly Pareto optimal} if there does not exist another point $\theta \in \R^n$ such that $F_i(\theta) < F_i(\theta^*)$ for all $i = 1,\dots, m$. The set of all weakly Pareto optimal points is the \emph{weak Pareto set}, which we denote by $P_w$ and the set $F(P_w)$ is the \emph{weak Pareto front}.
 \end{enumerate}
\label{def:Pareto_opt}
\end{mydef}

It is not straight-forward to design efficient optimization methods from the definitions above. To obtain more sophisticated algorithms we use first-order information and proximal point evaluations of the objective functions. In general, these methods do not compute Pareto optimal but Pareto critical points.

\begin{mydef}
A point $\theta^* \in \R^n$ is called \emph{Pareto critical} if there exist $\lambda \in \R^m$ with $\lambda_i \ge 0$ for all $i = 1,\dots, m$ and $\sum^m_{i=1}\lambda_i = 1$, satisfying 
\begin{align}
\label{eq:KKT}
    \sum^m_{i=1}\lambda_i \partial F_i(\theta^*) \ni 0,
\end{align}
where $\partial F_i(\theta^*)$ is the subdifferential.
\end{mydef}

\begin{figure}[h]
    \centering
    \begin{tikzpicture}[scale=0.4] 

        \draw[rotate=45, line width=0.6pt] (-0.95, -0.95) rectangle (0.95, 0.95)node[label={[label distance=-2.5cm, xshift=2.2cm]above:\textcolor{black}{$F_2(\theta) = \lVert \theta \rVert_1$}}] {};
        
        
        \draw[rotate=-45, line width=0.6pt] (-1.9, -1.9) rectangle (1.9, 1.9);
        
        
        \draw[rotate=-45, line width=0.6pt] (-2.85, -2.85) rectangle (2.85, 2.85);


        \draw[rotate=-45, line width=0.6pt] (-3.8, -3.8) rectangle (3.8, 3.8);

        \draw[->, line width=0.8pt] (-7, 0) -- (7, 0) node[right] {$\theta_2$};
        \draw[->, line width=0.8pt] (0, -6) -- (0, 8) node[above] {$\theta_1$};
    
        \begin{scope}[xslant=-0.4]
            
            \draw[line width=0.6pt] (-0.5, 3.5) circle (0.875);
            
            
            \draw[line width=0.6pt] (-0.5, 3.5) circle (1.75)  node[label={[above=1.2cm, xshift=-1.8cm]above:\textcolor{black}{$F_1(\theta) = \frac{1}{2}\lVert A \theta - b \rVert_2^2 $}}] {};
            
            
            \draw[line width=0.6pt] (-0.5, 3.5) circle (2.625) ;

            \draw[line width=0.6pt] (-0.5, 3.5) circle (3.5) ;
    
            \coordinate (ellipseIntersection) at (-0.5, 3.5);
            \coordinate (squareIntersection) at (0, 0);

            \draw[red, line width=2.0pt ] (ellipseIntersection) -- node[above right = 0.65cm, red] {$\left\lbrace \theta^*(\lambda) \, : \, \lambda \ge 0 \right\rbrace $} (0.8, 2) -- (squareIntersection);

        \end{scope}

        \draw[red, fill=red] (0,0) circle (.07);
        \draw[red, fill=red] (-1.9, 3.5) circle (.07);
        
    \end{tikzpicture}
    \caption{Contour plot of smooth objective functions $F_1(\theta) = \frac{1}{2}\|A\theta - b \|^2_2$ and non-smooth objective function $F_2(\theta) = \lVert \theta \rVert_1$ in black and the \emph{regularization path} $\theta^*(\lambda) \in \text{argmin}_{\theta \in \R^n} \frac{1}{2}\lVert A \theta - b \rVert_2^2 + \lambda\lVert \theta \rVert_1$ in red.}
    \label{fig:figure6}
\end{figure}
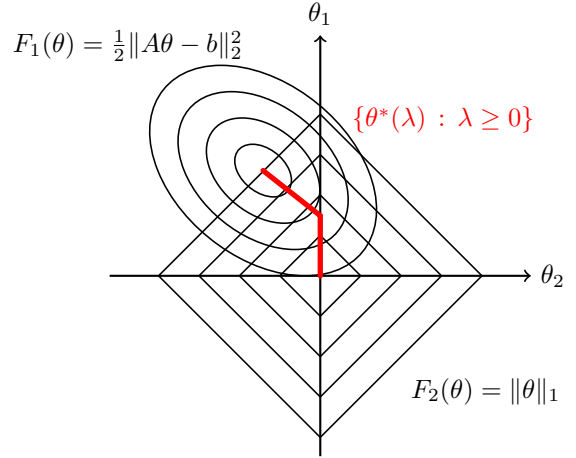

While all Pareto optimal points are Pareto critical, the reverse does not hold generally. If all objective functions are convex, then each Pareto critical point is weakly Pareto optimal. As an example, consider the LASSO problem, i.e., $\min_{x \in \R^n} \frac{1}{2}\|A\theta - b \|^2_2 + \lambda \| \theta \|_1$, with $A \in \R^{m \times n}$, $b \in \R^m$ and regularization parameter $\lambda \ge 0$. The LASSO problem is connected to the following \eqref{eq:MOP}  
\begin{align*}
    \min_{\theta \in \R^n} \left[ \begin{array}{c}
        \frac{1}{2}\|A\theta - b \|^2_2  \\
        \| \theta \|_1
    \end{array}\right].
\end{align*} 
For this problem the Pareto optimal points form a connecting path $\theta^*(\lambda)$, between the respective minimizers of the smooth objective, i.e., $\frac{1}{2}\|A\theta - b \|^2_2$ and the non-smooth objective, i.e., $\| \theta \|_1$, illustrated in \emph{Figure \ref{fig:figure6}}.

In this work, we consider two objective functions, namely the empirical loss and the $\ell^1$ norm of the neural network weights. The Pareto set connecting the individual minima (at least locally), is also known as the \emph{regularization path}. In the context of MOPs, we are looking for the Pareto set of
\begin{align}
    \min_{\theta\in \R^n} \left[ \begin{array}{c}
        \mathbb{E}_{(x,y) \sim \mathcal{D}}[\mathcal{L}(f(\theta,x),y)]  \\
        \frac{1}{n}\| \theta \|_1
    \end{array}\right],
    \label{eq:MoDNN}
    \tag{MoDNN}
\end{align}
where $(x,y) \in \mathcal{X}\times\mathcal{Y}$ is labeled data following a joint distribution $\mathcal{D}$, the function $f: \R^n \times \mathcal{X} \to \mathcal{Y}$ is a parameterized model and $\mathcal{L}(\cdot,\cdot)$ denotes a loss function. The second objective is the weighted $\ell^1$ norm $\frac{1}{n}\| \theta \|_1$ to ensure sparsity. Our goal is to solve \eqref{eq:MoDNN} to obtain the regularization path. However, this problem is challenging, as the $\ell^1$ norm is not differentiable. 

\begin{myremark}
   A common approach to solve the problem \eqref{eq:MOP} (including all sorts of regularization problems) is the use of the \emph{weighted sum method} using an additional hyperparameter $\lambda \in \R^m$, with $\lambda_i \ge 0$ for all $i = 1,\dots, m$ and $\sum_{i=1}^m \lambda_i = 1$. The weighted sum problem reads as
\begin{align}
      \min_{\theta\in \R^n}F(\theta) = \sum_{i = 1}^m \lambda_i F_i(\theta) 
        \label{ws}
   \end{align}
In the numerical experiments, we compare our continuation method with the weighted sum approach. 
\end{myremark}

\subsection{Multiobjective proximal gradient method}

Given functions of the form $F_i = f_i + g_i$, such that $f_i: \R^n \to \R$ is smooth, and $g_i:\R^n \to \R$ is convex and non-smooth with computable proximal operator $\prox_{g}$, from problem \eqref{eq:MoDNN} we have $F_1(\theta) = f_1(\theta) = \mathbb{E}_{(x,y) \sim \mathcal{D}}[\mathcal{L}(f(\theta,x),y)]$ and $F_2(\theta) = g_2(\theta) = \frac{1}{n}\| \theta \|_1$. The proximal operator $\textnormal{prox}_{\frac{1}{n}\| \cdot \|_1}(\theta)$ has a simple closed form. This allows for an efficient implementation of Algorithm \ref{algo:example1}, which yields a single Pareto critical point for MOPs with objectives of this type.
In \cite{Tanabe2019} it is shown that Algorithm \ref{algo:example1} converges to Pareto critical points.

\begin{mydef}[Proximal operator]
    Given a convex function $g : \R^n \to \R $, the proximal operator is
\begin{align*}   
 \hspace{5mm}\operatorname{prox}_{\substack{g\\}}(\theta) =  \underset{\phi \in \R^n}{\arg\min} \left\{ g(\phi) + \frac{1}{2} \|\phi - \theta\|_{2}^2 \right\}.
\end{align*}
\label{definition4}
\end{mydef}

\begin{algorithm} 
	\caption{Multiobjective Proximal Gradient \cite{Tanabe2019} }
        \textbf{Input} Initialize $k=0$.  \\
        \textbf{Parameter} $\theta^0 \in \R^n$, step size $h > 0$.\\
        \textbf{Output}: $\theta^k$ 
        
	\begin{algorithmic}[1] 
		\STATE \textbf{Compute} the descent direction $d^k$ by solving  
            \begin{align*}
                 d^k = \underset{d \in \R^n}{\arg\min}\left\{\psi_{\theta^k}(d) + \frac{1}{2h} \|d\|_2^2 \right\}
            \end{align*}
            where
            \begin{align*}
                \psi_{\theta}(d) = \underset{i=1,\dots,m}{\max} \left\{ \nabla f_i(\theta)^Td + g_i(\theta + d) -g_i(\theta)\right\}  
            \end{align*}
         \STATE \textbf{if} $d^k=0$, \textbf{STOP}   
        \STATE \textbf{Update} $\theta^{k+1} = \theta^k + d^k$
            \STATE \textbf{Set} $k = k+1$ and go to step 1
	\end{algorithmic} 
	\label{algo:example1}
\end{algorithm}	

\begin{myremark}
Proximal gradient iterations are “forward-backward” iterations, "forward" referring to the gradient step and "backward" referring to the proximal step. The proximal gradient method is designed for problems where the objective functions include a smooth and a non-smooth component, which is suitable for optimization with a sparsity-promoting regularization term.
\end{myremark}


\begin{theorem}\cite{Tanabe2019}
Let $f_i$ be convex with $L$-Lipschitz continuous gradients and let $g_i$ be proper, convex and lower semi-continuous for all $i=1, \dots ,m$ with step size $h \leq \frac{2}{L}$. Then, every accumulation point of the sequence $\{\theta^k \}$ computed by \textit{Algorithm \ref{algo:example1}} is Pareto critical.
In addition, if the level set of one $F_i$ is bounded, the sequence $\{\theta^k \}$ has at least one accumulation point.
\end{theorem}

    

\subsection{A predictor-corrector method}

\begin{figure}[b]
  \centering
 \includegraphics[width=0.7\columnwidth]{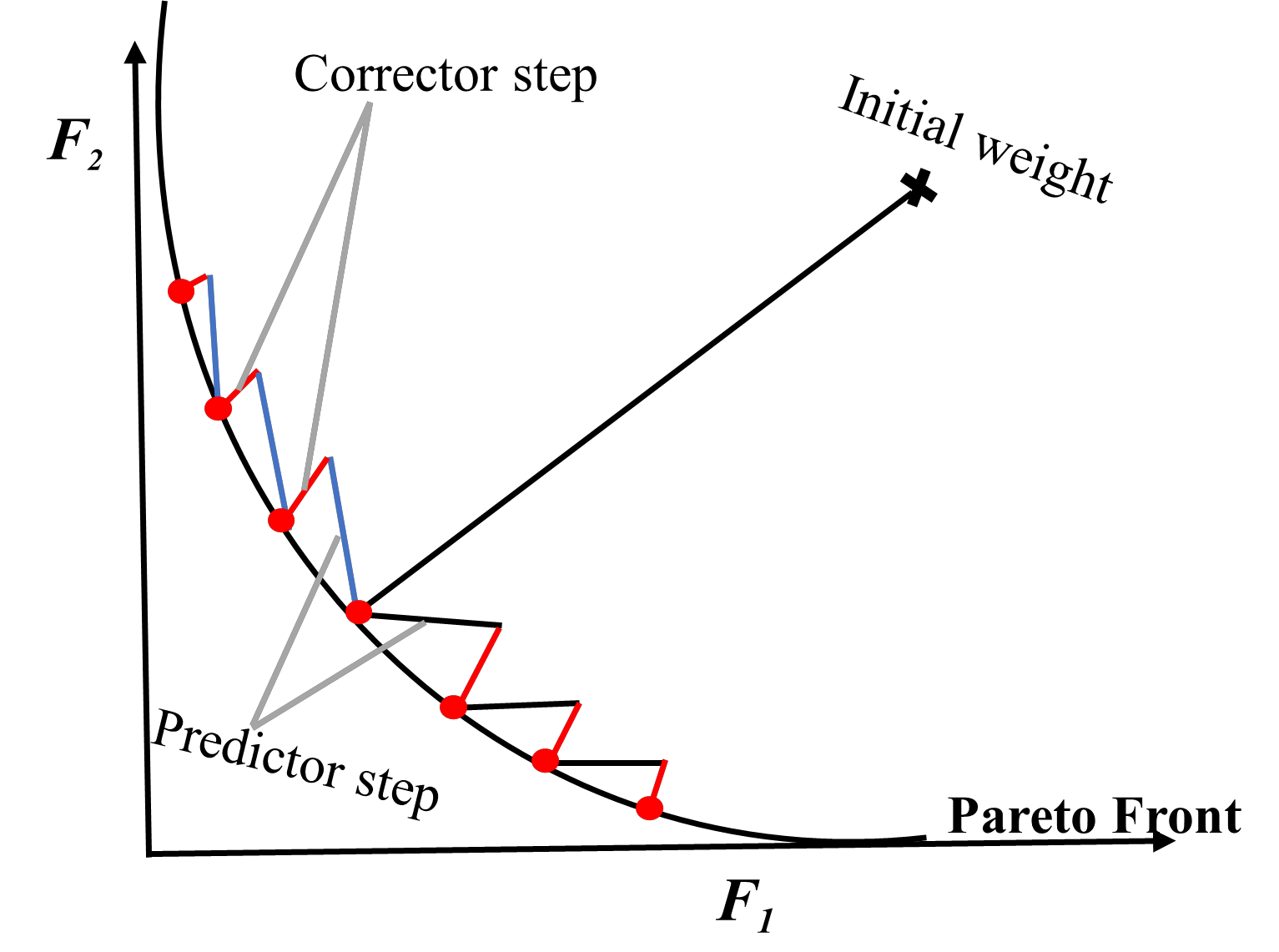}
  \caption{Sketch of the continuation method; The predictor steps are shown in black and blue for Eqs.\ \eqref{eq:gradient} and \eqref{eq:shrinkage} respectively. The corrector step is shown in the red lines and the red dots are the Pareto optimal points computed by the predictor-corrector method.}
  \label{img:cont}
\end{figure}

This section describes our continuation approach to compute the entire Pareto front of problem \eqref{eq:MoDNN}. Fig.\ \ref{img:cont} shows an exemplary illustration of the Pareto front approximated by a finite set of points that are computed by consecutive predictor and corrector steps. We start by finding an initial point $\theta^0$ on the Pareto front using Algorithm \ref{algo:example1} given that it always converges to a global Pareto critical point \cite{Tanabe2019}, then we proceed along the front. This is done by first performing a predictor step in a suitable direction which takes us close to a different part of the front \cite{Hillermeier2001, Bieker2022}, then we need to perform a consecutive corrector step that takes us back to the front.

\vspace{2mm}
\paragraph{Predictor step} 
Depending on the direction we want to proceed in, we perform a predictor step simply by performing a gradient step or proximal point operator step: 
\begin{align}
 \theta^{k+1} &= \theta^{k} -\eta \nabla_{\theta} \mathbb{E}_{(x,y) \sim \mathcal{D}}[\mathcal{L}(f(\theta^k,x),y)],
\label{eq:gradient} \\
  \mbox{or}\qquad\theta^{k+1} &= \operatorname{prox}_{\substack{{\eta \| \cdot \|_1}\\}}(\theta^{k}).
\label{eq:shrinkage}
\end{align}

Equation \eqref{eq:gradient} is the gradient step for the loss objective function, i.e., ``move left" in Fig.\ \ref{img:cont} and equation \eqref{eq:shrinkage} represents the shrinkage performed on the $\ell^1$ norm, i.e., ``move down" in Fig.\ \ref{img:cont}.
Note that this is a deviation from the classical concept of continuation methods as described previously, where we compute the tangent space of a manifold. However, due to the non-smoothness, the Pareto set of our problem does not possess such a manifold structure, which means that we cannot rely on the tangent space. Nevertheless, this is not necessarily an issue, as the standard approach would require Hessian information, which is too expensive in high-dimensional problems.
The approach presented here is significantly more efficient, even though it may come at the cost that the predictor step is sub-optimal. Despite this fact, we found in our numerical experiments that our predictor nevertheless leads to points close enough for the corrector to converge in a small number of iterations. 

\vspace{2mm}
\paragraph{Corrector step} For the corrector step, we apply the multiobjective proximal gradient method (MPG) described in Algorithm \ref{algo:example1}. As our goal is to converge to a point on the Pareto front, this step is identical for both predictor directions. 

The method is summarized in Algorithm \ref{algo:example2} for both directions, which only differ in terms of line 4.

\begin{algorithm}
\caption{Predictor-Corrector Method}
\label{algo:example2}
\textbf{Input}: Number of predictor-corrector runs $N_{cont}$, $P \gets \{ \}$ \\
\textbf{Parameter}: Initial parameter $\theta^0 \in \R^n$, learning rate $\eta$\\
\textbf{Output}: $P$ approximate Pareto set 

\begin{algorithmic}[1] 
\STATE \textbf{Compute} update to $\theta^0$ using \textit{Algorithm \ref{algo:example1}} with initial value $\theta^0$ and step size $h = \eta$.
\STATE \textbf{Update} $P = P \cup \{ \theta^0\}$
\FOR{$n = 1: (N_{cont} - 1)$}
\STATE \textbf{Compute} predictor $\theta_p^{n}$ by performing a predictor step from $\theta^{n-1}$ with equation\ \eqref{eq:gradient} or \eqref{eq:shrinkage}
\STATE \textbf{Compute} corrector $\theta^{n}$ by applying \textit{Algorithm \ref{algo:example1}} with initial value $\theta_p^{n}$ and step size $h = \eta$.
 \STATE \textbf{Update} $P = P \cup \{ \theta^n\}$
\ENDFOR 
\end{algorithmic} 
\end{algorithm}

\section{Numerical experiments}
\label{Num}
In this section, we present numerical experiments for our algorithm and the resulting improvements in comparison to the much more widely used weighted sum (WS) approach in equation \eqref{ws} and as an additional experiment, we include results from the evolutionary algorithm (EA) specifically the non-dominated sorting genetic algorithm (NSGA-II) which has been claimed to perform well by finding better spread nondominated Pareto optimal points \cite{Deb2002} for MOPs. We show its performance in low and high-dimensional cases in comparison to the predictor-corrector method (CM) and the WS method.

In real-world problems, uncertainties and unknowns exist. Our previous approach of using the full gradient on every optimization step oftentimes makes it computationally expensive, as well as unable to handle uncertainties \cite{Mitrevski2020}. A stochastic approach is included in our algorithm to take care of these limitations. The stochasticity is applied by computing the gradient on mini-batches of the data.

\subsection{Experimental settings}
\label{ExpSet}

To evaluate our algorithm, we first perform experiments on the Iris dataset \cite{iris1988}. Although this dataset is by now quite outdated, it allows us to study the deterministic case in detail. We then extend our experiments to the well-known MNIST dataset \cite{dengmnist2012} and the CIFAR10 dataset \cite{krizhevsky2014cifar} using the stochastic gradient approach. The Iris dataset contains 150 instances, each of which has 4 features and belongs to one of 3 classes. The MNIST dataset contains 70000 images, each having 784 features and belonging to one of 10 classes. The CIFAR10 dataset contains 60000 images of size 32x32 pixels with 3 color channels and contains images belonging to one of 10 classes. We split all datasets into training and testing sets in an 80--20 ratio.

\begin{figure}[h]
\centering
\includegraphics[width=0.97\linewidth]{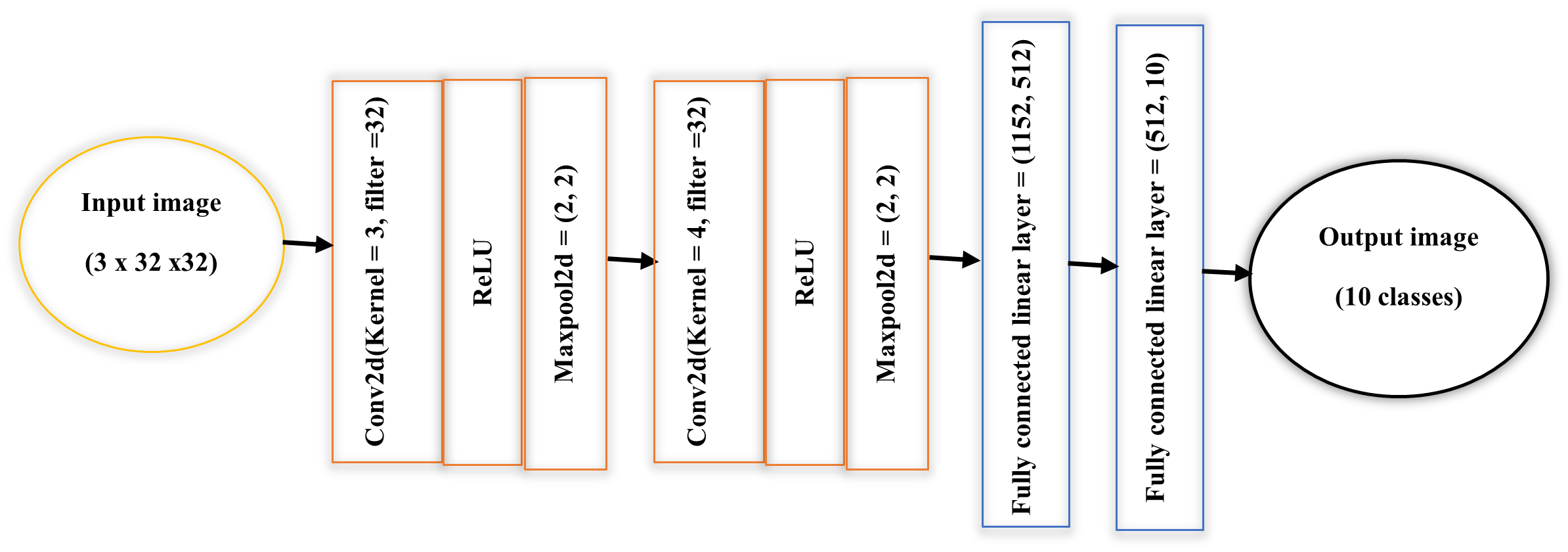}
\caption{The neural network architecture used for testing the CM and WS methods on CIFAR10 dataset.}
\label{fig:cnnCifar}
\end{figure}

We use a dense linear neural network architecture with two hidden layers for both the Iris and MNIST datasets (4 neurons and 20 neurons per hidden layer, respectively), with ReLU activation functions for both layers. For the CIFAR10 dataset, two fully connected linear layers after two convolutional layers are used as shown in Fig. \ref{fig:cnnCifar}. Cross-entropy is used as the loss function.

The experiments for the Iris and MNIST datasets are carried out on a machine with 2.10 GHz 12th Gen Intel(R) Core(TM) i7-1260P CPU and 32 GB memory, using Python 3.8.8 while the CIFAR10 experiment is carried out on a computer cluster with a NVIDIA A100 GPU, 64 GB Ram and a 32-Core AMD CPU with 2.7GHz using Python 3.11.5. The source code is available at \url{https://github.com/aamakor/continuation-method}.

\subsection{Experimental Procedure}
\begin{figure}
\centering
\includegraphics[width=0.7\linewidth]{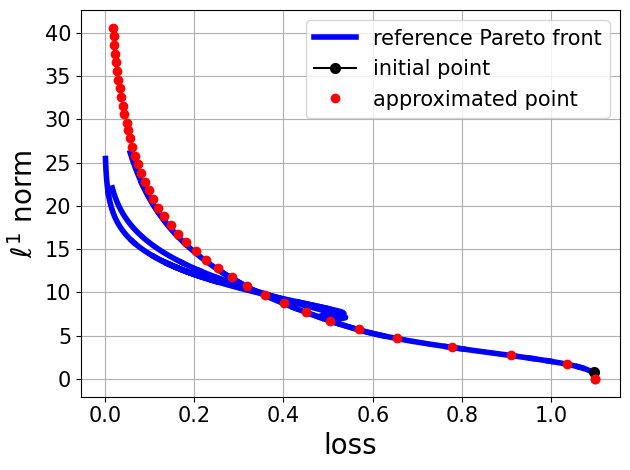}
\caption{Pareto front approximation for the Iris dataset using \emph{Algorithm 2} (red symbols) versus the reference Pareto front in ``blue" (computed using the same algorithm with very small step sizes and many different initial conditions) with unscaled $\ell^1$ norm.}
\label{fig:sub2}
\end{figure}

First, we illustrate the behavior of our algorithm by studying the Iris dataset in a deterministic setting. To obtain a baseline, we executed Algorithm \ref{algo:example2} using very small step sizes $h$. Interestingly, the Pareto set and front consist of multiple components, which we were only able to find by repeated application of the continuation method with random initial conditions (multi-start). The resulting solutions are shown in blue in Fig.\ \ref{fig:sub2}, where three connected components of the Pareto critical points are clearly visible. As this approach is much too expensive for a realistic learning setting (the calculations took close to a day for this simple problem), we compare this to a more realistic setting in terms of step sizes. The result is shown via the red symbols which were computed within seconds. Motivated by our initial statement on more sustainable network architectures, we have initialized our neural network with all weights being very close to zero (the black ``$\bullet$'' in Fig.\ \ref{fig:sub2}) since the network is small and then we proceed along the front towards a less sparse but more accurate architecture\footnote{Due to symmetries, an initialization with all zeros poses problems in terms of which weights to activate, etc., see \cite{Bieker2022} for details.} using the predictor corrector steps described in Algorithm \ref{algo:example2}.

As we do not need to compute every neural network parametrization from scratch, but use our predictor-corrector scheme, the calculation of each individual point on the front is much less time-consuming than classical DNN training. Moreover, computing the slope of the front from consecutive points allows for the online detection of relevant regions. Very small or very large values for the slope indicate that small changes in one objective lead to a large improvement in the other one, which is usually not of great interest in applications. Moreover, this can be used as an early stopping indicator to avoid overfitting.


\subsection{Results}
\label{Res}
Having established a baseline as illustrated in \ref{fig:sub2} using the Iris dataset for a simple neural network architecture, we extend the predictor-corrector algorithm to more high-dimensional cases ie., MNIST and CIFAR10 datasets and compare the results against the weighted sum algorithm in equation\ \eqref{ws}, where $\lambda$ is now the weighting parameter chosen equidistantly on the interval $[0,1]$. For the gradient descent optimization, in both the predictor step of our algorithm and the weighted sum, the ADAM optimizer described in \cite{Adam2015} has been used.

\vspace{2mm}
\subsubsection{MNIST}
\begin{figure}[t]
\begin{subfigure}{.5\textwidth}
  \centering
  \includegraphics[width= 0.6\linewidth]{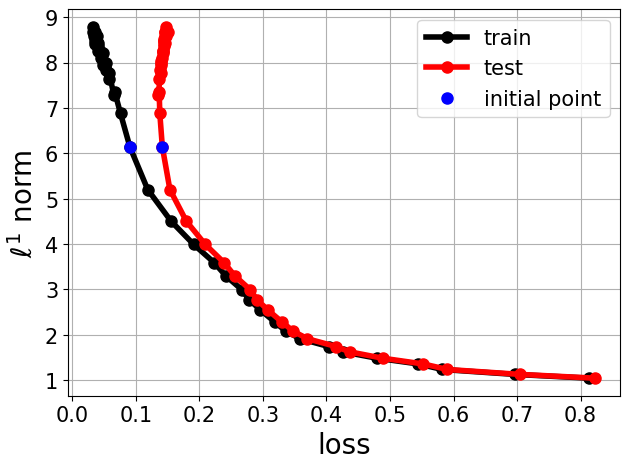}
  \caption{}
  \label{fig:sub3}
\end{subfigure}
\begin{subfigure}{.5\textwidth}
  \centering
  \includegraphics[width=0.6\linewidth]{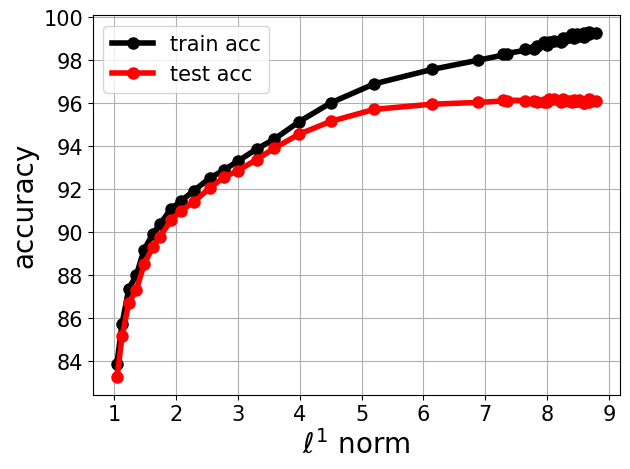}
  \caption{}
  \label{fig:sub4}
\end{subfigure}
\caption{(\subref{fig:sub3}) The Pareto front for the MNIST dataset (in black), where the initial point is shown in blue. The red curve shows the performance on the validation set. Non-sparse networks clearly tend to overfit. (\subref{fig:sub4}) The prediction accuracy versus $\ell^1$ norm, where the overfitting regime becomes apparent once more.}
\label{fig:test1}
\end{figure}

Motivated by the promising results in the deterministic setting, we now study the MNIST dataset in a stochastic setting, i.e., using mini-batches for the loss function and a neural network with $16,330$ parameters as described previously in \ref{ExpSet}. First, we obtain the initial point on the Pareto front after performing 500 iterations, using Algorithm \ref{algo:example1}. The remaining points on the Pareto front are computed by using small consecutive predictor corrector steps. For the subsequent 43 points of the Pareto front found after the initial point, we used 7 iterations for the predictor steps and 20 iterations for the corrector steps. In total, we computed 44 points, i.e., $N_{cont}=44$. Figures \ref{fig:test1} (\subref{fig:sub3}) and (\subref{fig:sub4}) show the Pareto front and accuracy versus the $\ell^1$ norm, respectively. In this setting unlike in the Iris dataset, we did not set our neural network weights to zeros rather we started by finding a point in the middle of the front in blue and then applied the continuation method twice (once in each direction, i.e., loss and $\ell^1$ norm). As indicated in the plots, we observe overfitting for the non-sparse architectures, which indicates that we do not necessarily have to pursue the regularization path until the end, but we can stop once the slope of the Pareto front becomes too steep. This provides an alternative training procedure for DNNs where in contrast to pruning we start sparse and then get less sparse only as long as we don't run into overfitting.

\begin{figure}
\begin{subfigure}{.5\textwidth}
  \centering
  \includegraphics[width=0.65\linewidth]{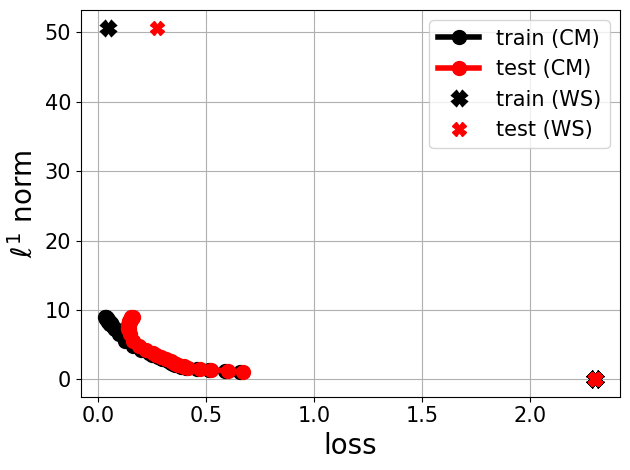}
  \caption{}
  \label{fig:sub5}
\end{subfigure}
\begin{subfigure}{.5\textwidth}
  \centering
  \includegraphics[width=0.65\linewidth]{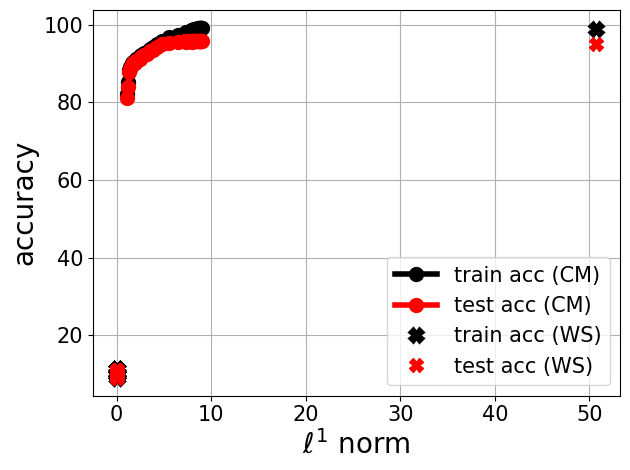}
  \caption{}
  \label{fig:sub6}
\end{subfigure}
\caption{Comparison between the predictor-corrector (CM) approach and weighted sum (WS) approach in the MNIST dataset. The figures show the same plots as Fig.\ \ref{fig:test1} but include the WS solutions. A clustering around the sparse and non-regularized solutions is evident, even though we have used equidistantly distributed weights $\lambda$.}
\label{fig:test2}
\end{figure}

For comparison, we compute the Pareto front using the weighted sum approach. For the WS, we aim also, to find 44 points on the Pareto front as in the case of the CM, i.e., $N_\lambda = 44$. As mentioned previously, we start by selecting 44 equidistantly distributed weights $\lambda$, where $\lambda \in [0, 1]$ and using ADAM optimizer (known for its fast convergence) for the gradient descent. Each Pareto point computed took 200 iterations and these 200 iterations (epochs) were repeated 44 times. In total, from the computation time as shown in TABLE \ref{table:resultsMM}, we observe that the WS method used 5 times as many epochs as the continuation method (\emph{Algorithm \ref{algo:example2}}) and therefore 5 times as many forward and backward passes. Also, the WS approach results in clustering around the unregularized and very sparse solutions, respectively as illustrated in Fig.\ \ref{fig:test2} where the comparison between the CM and the WS approaches is seen. This behaviour of the WS is in accordance with previous findings from \cite{Das1997, Marler2009, Bieker2022}. In contrast, we obtain a similar performance with a much sparser architecture using continuation. This shows the superiority of the continuation method and also highlights the randomness associated with simple regularization approaches when not performing appropriate hyperparameter tuning.

\vspace{2mm}
\subsubsection{CIFAR10}
\begin{figure}
\begin{subfigure}{.5\textwidth}
  \centering
  \includegraphics[width=0.65\linewidth]{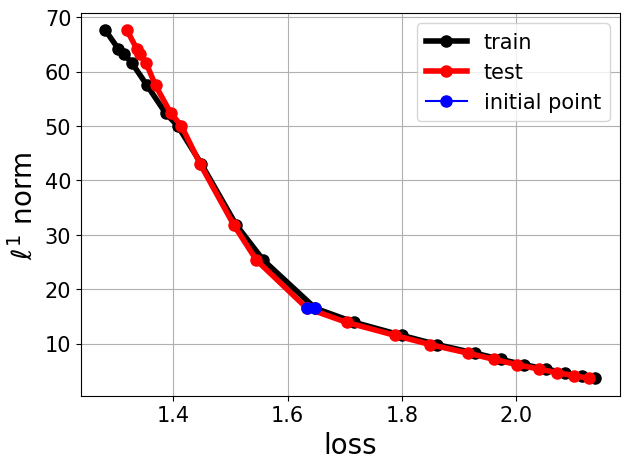}
  \caption{}
  \label{fig:sub7}
\end{subfigure}
\begin{subfigure}{.5\textwidth}
  \centering
  \includegraphics[width=0.65\linewidth]{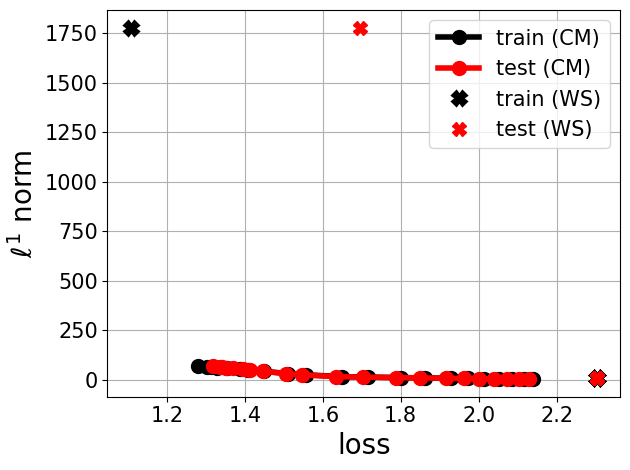}
  \caption{}
  \label{fig:sub8}
\end{subfigure}
\caption{ (\subref{fig:sub7}) The Pareto front for the CIFAR10 dataset (in black), where the initial point is shown in blue. The red curve shows the performance on the test set. Non-sparse networks clearly tend to overfit. (\subref{fig:sub8}) The same plots as (\subref{fig:sub7}) but including the WS solutions. A clustering around the sparse and non-regularized solutions is evident in the WS method plot, even though we have used equidistantly distributed weights $\lambda$.}
\label{fig:test3}
\end{figure}

To further test the performance of our algorithm we extend to a higher-dimensional DNN with complex architecture as shown in Fig.\ \ref{fig:cnnCifar} for the CIFAR10 dataset. This neural network architecture has  $4,742,546$ parameters as presented in TABLE \emph{\ref{table:resultsMM}}. We observe that our method provides a well-spread points for the Pareto front. Fig.\ \ref{fig:sub7} shows the initial point obtained after $2000$ iterations and the subsequent points obtained using the predictor step ($7$ iterations) and corrector step ($25$ iterations) for a total of $N_{cont}= 20$ points on the front.

Again, for comparison, we compute the same number of points on the Pareto front using the WS approach, i.e., $N_\lambda = 20$ where $\lambda = 20 \ \textnormal{and} \ \lambda \in [0, 1]$. Each points on the front takes  $118$ epochs to compute. Details of the computation time are shown in TABLE \ref{table:resultsMM}. Fig.\ \ref{fig:sub8} shows the Pareto front comparison of the CM with the WS for the CIFAR10 dataset and it can be observed once again---as for MNIST---that the WS yields clusters at the minima of the individual objectives. Furthermore, we see that the non-sparse solution exhibits overfitting. 
Hence, CM provides a good trade-off between sparsity and loss in DNNs, not just for a very sparse DNN as seen in the MNIST and CIFAR10 cases. Hence, we obtain a well-structured regularization path for nonlinear high-dimensional DNNs.

We note, that several iterations should be used until convergence when obtaining the initial point on the Pareto front for the predictor-corrector method. Once the initial point is identified, which is comparable to the complexity of the scalarized problem, the follow-up computations require fewer iterations and are much cheaper, when compared with the WS method.

\begin{table} 
\begin{adjustbox}{width=\linewidth,center}
  \centering
  \renewcommand{\arraystretch}{1} 
  \begin{tabular}{*{10}{>{\centering\arraybackslash}p{1.7cm}}}
    \Xhline{1.5pt} 
    \multirow{2}{*}{Data (parameters)} & \multirow{2}{*}{Method} & \multirow{2}{*}{\makecell{Num \\ iterations}} & \multirow{2}{*}{\makecell{Computation\\ time (sec)}} & \multicolumn{2}{c}{Accuracy} \\
    & & & & \makecell{Training \\ Accuracy} & \makecell{Testing \\ Accuracy} \\
    \Xhline{1.5pt} 
    \multirow{5}{*}{MNIST ($16,330$)} & \multirow{2}{*}{WS} &  \multirow{2}{*}{$49.28e6$} & \multirow{2}{*}{$1140$} & \multirow{2}{*}{\bfseries 98.83$\%$} & \multirow{2}{*}{$94.94\%$} \\ 
    & &  &  &  &  \\
    \cmidrule(l{17pt}r{0pt}){2-6}
    & \multirow{2}{*}{CM} & \multirow{2}{*}{$9.3e6$ }& \multirow{2}{*}{$329$} & \multirow{2}{*}{$97.92\%$} & \multirow{2}{*}{\bfseries 95.66$\%$} \\
& &  &  &  &  \\
    \Xhline{1.5pt} 
    \multirow{5}{*}{CIFAR10 ($4,742,546$)} & \multirow{2}{*}{WS} &  \multirow{2}{*}{$1.77e6$} & \multirow{2}{*}{$575$} & \multirow{2}{*}{\bfseries 60.95$\%$} & \multirow{2}{*}{$41.63\%$} \\ 
    & &  &  &  &  \\
    \cmidrule(l{25pt}r{1pt}){2-6} 
    & \multirow{2}{*}{CM} & \multirow{2}{*}{$1.95e6$ }& \multirow{2}{*}{$1193$} & \multirow{2}{*}{$54.45\%$} & \multirow{2}{*}{\bfseries 52.80$\%$} \\
& &  &  &  &  \\
    \Xhline{1.5pt} 
  \end{tabular}
   \end{adjustbox}
   \caption{Settings and Results on MNIST and CIFAR10 dataset. For each metric, the best performance per architecture is in bold.}
    \label{table:resultsMM}
\end{table}

\subsection{Comparison with Evolutionary Algorithms}
\label{AddE}

\begin{figure}
\begin{subfigure}{.5\textwidth}
  \centering
  \includegraphics[width=0.65\linewidth]{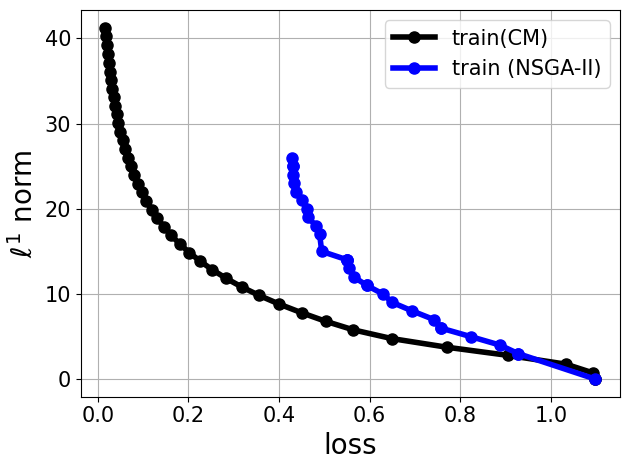}
  \caption{}
  \label{fig:sub29}
\end{subfigure}
\begin{subfigure}{.5\textwidth}
  \centering
  \includegraphics[width=0.65\linewidth]{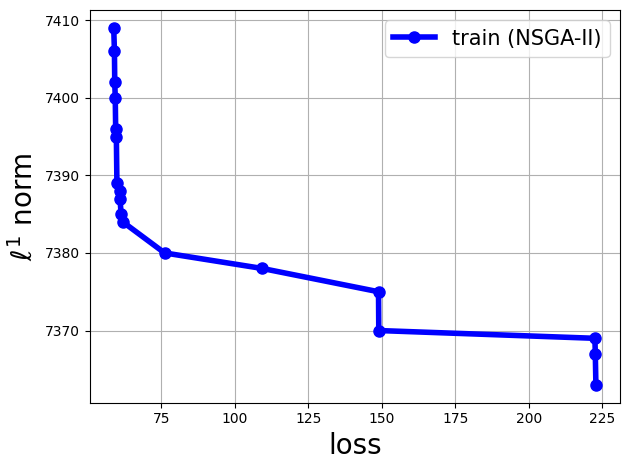}
  \caption{}
  \label{fig:sub30}
\end{subfigure}
\begin{subfigure}{.5\textwidth}
  \centering
  \includegraphics[width=0.65\linewidth]{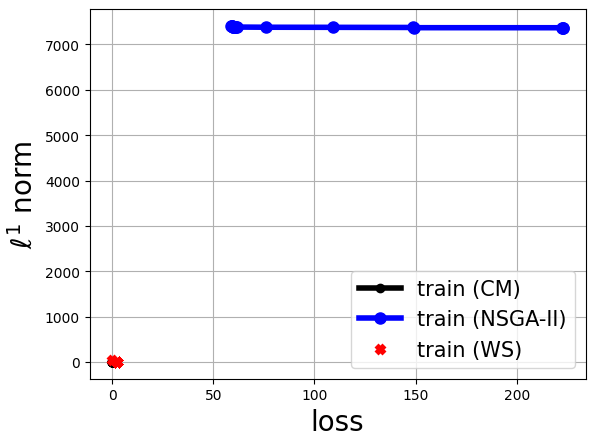}
  \caption{}
  \label{fig:sub31}
\end{subfigure}
\caption{ (\subref{fig:sub29}) The Pareto front for the Iris dataset in the deterministic setting, where NSGA-II is in blue and the CM is shown in black. (\subref{fig:sub30}) The Pareto front for the MNIST train dataset, NSGA-II. From the scale of the NSGA-II plot, it is evident that the NSGA-II does not converge for high dimensions. (\subref{fig:sub31}) shows how far the Pareto front for NSGA-II is, from the CM and WS method for MNIST dataset.}
\label{fig:test15}
\end{figure}

As an additional experiment, in order to solidify our assertion on the efficiency of our algorithm on handling high-dimensional problems in comparison to other methods, we have also attempted to compute the Pareto fronts for the Iris dataset (deterministic) and MNIST dataset (stochastic) using the evolutionary algorithm (EA). We use one of the most popular and very efficient EA as describe by \cite{Rahimi2023}, the NSGA-11. Using a population size of $100$ with $100$ generations and a Two point crossover (having tried a crossover probability $0.9$ and distribution index 10) -- details present in the code, we observe that for the Iris dataset, the NSGA-11 algorithm found $30$ points that were dominated in terms of the training dataset and have a test accuracy of $93.33\%$ on the test set. For the MNIST dataset with $16,330$ parameters, the NSGA-11 algorithm fails to converge and also produces $18$ points that are heavily dominated with a computational time of approximately $810\,\text{sec}$ for the train set and a test accuracy of only $17.80\%$. Fig.\ \ref{fig:test15} illustrates the different fronts obtained when the evolutionary algorithm, NSGA-11 is used. Fig.\ \ref{fig:sub29} shows the front obtained from the NSGA-11 for a low-dimensional problem (Iris - $55$ parameters), its behaviour in comparison to the predictor-corrector step of the CM algorithm.
Fig.\ \ref{fig:sub30} and \ref{fig:sub31} shows how poor the NSGA-11 scales for high-dimensional multiobjective problems regardless of being more computationally expensive as compared to the CM method. This shows the superiority of our approach and its ability to perform well in high-dimensional cases.

\section{Conclusion}
We have presented an extension of regularization paths from linear models to high-dimensional nonlinear deep learning models. This was achieved by extending well-known continuation methods from multiobjective optimization to non-smooth problems and by introducing more efficient predictor and corrector steps. The resulting algorithm shows a performance that is suitable for high-dimensional learning problems. 

Moreover, we have demonstrated that starting with sparse models can help to avoid overfitting, e.g., the MNIST and CIFAR10 where high sparsity ($\ell^1$ norm) results in small loss as illustrated in the plots and also significantly reduce the size of neural network models. Due to the small training effort of consecutive points on the Pareto front, this presents an alternative, structured way to deep neural network training, which pursues the opposite direction than pruning methods do. Starting sparse, we increase the number of weights only as long as we do not obtain a Pareto front which is too steep, as this suggests overfitting.

For future work, we will consider more objectives. Our approach works very well on extremely high-dimensional problems, e.g., CIFAR-10 with two objectives, where the Pareto front is a line. Extending our work to cases with more than two objective functions will require significant extensions to the concept of adaptive Pareto exploration, e.g., \cite{Schtze2019}. Since the Pareto set becomes a higher-dimensional object, it will no longer be useful to compute the entire set but steer along desired directions in order to meet a decision maker's desired trade-off.

\section*{Acknowledgements}
This work was supported by the German Federal Ministry of Education and Research (BMBF) funded AI junior research group “Multicriteria Machine Learning".



\bibliography{IEEEtran.bib}
\bibliographystyle{IEEEtran}

\end{document}